\documentclass{article}
\usepackage{ijcai22}

\usepackage{times}

\usepackage{soul}
\usepackage{url}
\usepackage[hidelinks]{hyperref}
\usepackage[utf8]{inputenc}
\usepackage[small]{caption}
\usepackage{graphicx}
\usepackage{amsmath}
\usepackage{booktabs}
\usepackage{color}
\usepackage{graphics} % for pdf, bitmapped graphics files
\usepackage{epsfig} % for postscript graphics files
\usepackage{mathptmx} % assumes new font selection scheme installed
\usepackage{times} % assumes new font selection scheme installed
\usepackage{amsmath} % assumes amsmath package installed
\usepackage{amssymb}  % assumes amsmath package installed
\usepackage{multirow}
\usepackage{caption}
\usepackage{subcaption}
\usepackage{graphicx} % omit 'demo' for real document
\usepackage{hyperref}
\usepackage{color}
\usepackage{makecell}
\usepackage{colortbl}
\usepackage{caption}
\definecolor{mygray}{gray}{.9}
\usepackage{verbatim}
\title{\LARGE \bf
Self-supervised 3D Semantic Representation Learning for Vision-and-Language Navigation
}

\author{
Sinan Tan$^1$\and
Mengmeng Ge$^1$\footnote{Equal Contribution.}\and
Di Guo$^1$\and
Huaping Liu$^1$\footnote{Corresponding Author.}\And
Fuchun Sun$^1$\\
\affiliations
$^1$Department of Computer Science and Technology, Tsinghua University, China\\
}

\begin{document}

\maketitle

\begin{abstract}

In the Vision-and-Language Navigation task, the embodied agent follows linguistic instructions and navigates to a specific goal. It is important in many practical scenarios and has attracted extensive attention from both computer vision and robotics communities. However, most existing works only use RGB images but neglect the 3D semantic information of the scene. To this end, we develop a novel self-supervised training framework to encode the voxel-level 3D semantic reconstruction into a 3D semantic representation. Specifically, a region query task is designed as the pretext task, which predicts the presence or absence of objects of a particular class in a specific 3D region. Then, we construct an LSTM-based navigation model and train it with the proposed 3D semantic representations and BERT language features on vision-language pairs. Experiments show that the proposed approach achieves success rates of 68\% and 66\% on the validation unseen and test unseen splits of the R2R dataset respectively, which are superior to most of RGB-based methods utilizing vision-language transformers.

\end{abstract}

\section{Introduction}

In the Vision-and-Language Navigation (VLN)\cite{vln} task, the visual perception, language understanding, and decision making of the embodied agent are combined as a loop, enabling the agent to navigate in the environment following the instruction. There are many related work\cite{cvdn,reverie} emerging in the fields of computer vision and robotics.
Most existing works use the monocular and panoramic RGB images\cite{sf} as the input and encode them to various features such as Faster R-CNN detector region features \cite{prevalent}, entity relation graph features \cite{relgraph}.

Although the performance on the typical VLN benchmark has increased dramatically with the RGB observation as the input, the depth information which can be obtained by the embodied agent is neglected. With only the RGB information, many semantic-irrelevant texture details are inevitably introduced into the training procedure, resulting in the model suffering from the overfitting problem, and the trained agent will not have sufficient capability of 3D environment perception. To tackle this problem, \cite{sun2021depth} proposes a network using depth features to normalize RGB features, while only a marginal performance improvement is observed. It may be because the joint learning of vision, language, and action modules actually requires more high-level 3D semantic information to overcome the overfitting problem. One possible approach is to introduce the 3D semantic information into the VLN task, while it is also challenging. Although the 3D semantic reconstructed can be obtained from RGB-D observations, in the context of the VLN tasks, the limited annotated data makes the encoding of the 3D semantic features difficult to learn, let alone the joint learning of the 3D semantic encoder and the navigation model.

% How to leverage the 3D semantic information in the VLN task is still an unresolved problem. 

%TODO DETR 
To solve the above challenge, we develop a novel self-supervised 3D semantic representation learning framework for the vision-and-language navigation task in this work. Inspired by the DETR detector, a region query task is designed as the pretext task, which expects the model to answer whether a particular class object is presented in a specific 3D region of the scene. Based on this mechanism, the model can learn to encode the 3D semantic reconstruction into a meaningful representation, which can be used for navigation tasks. Experiments show that the proposed 3D semantic representation improves performance compared with traditional panoramic RGB features.

% \cite{carion2020end}
    \begin{figure*}[!htbp]
    \centering
    \includegraphics[width=0.9\linewidth]{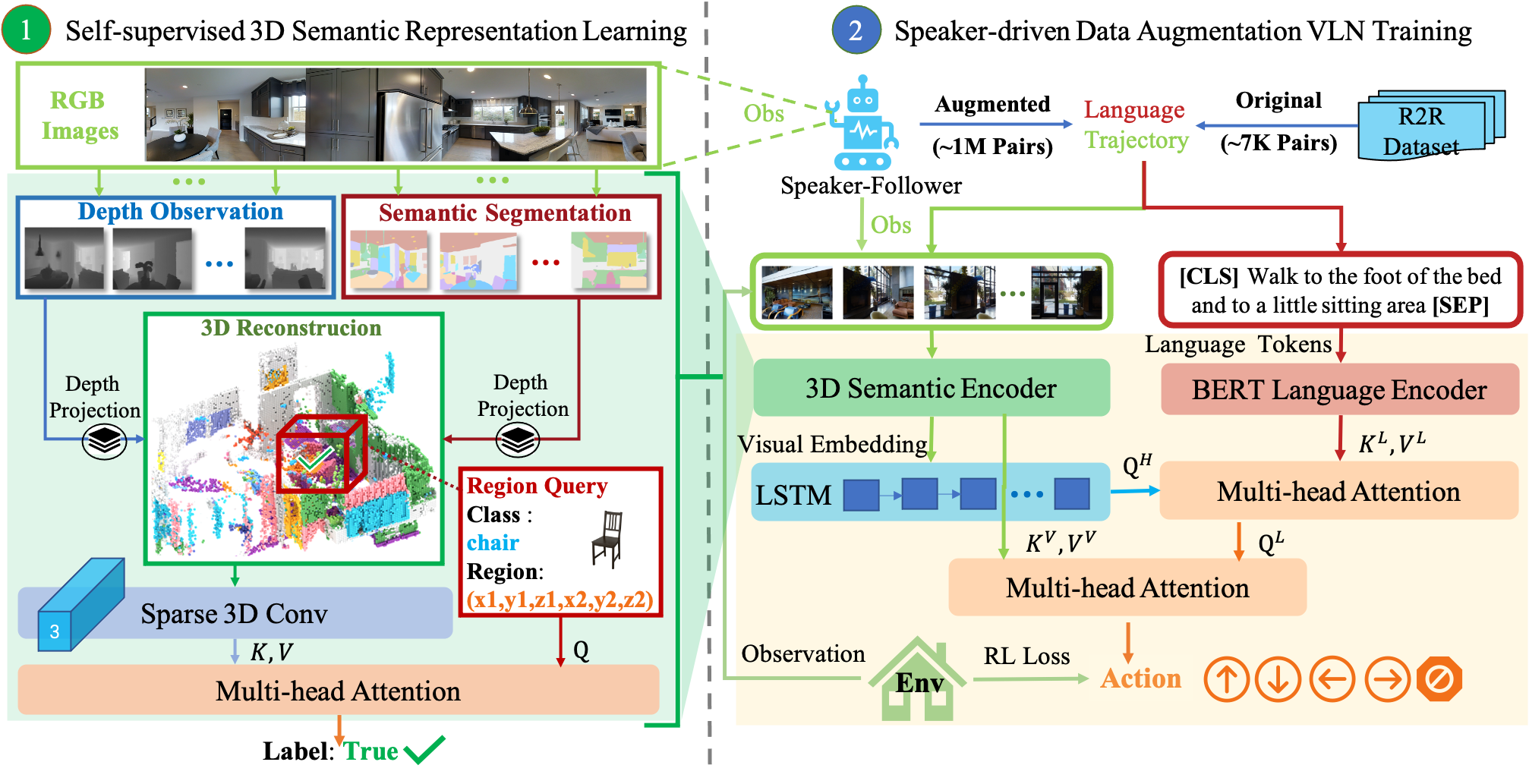}
    \caption{Our proposed method for the vision-and-language navigation task. Firstly, in order to encode the local 3D semantic information by self-supervised presentation learning, segmentation of RGB observations is combined with depth information to reconstruct 3D by projection, and then the 3D visual semantic encoder is trained in a self-supervised approach with the query task of regional object category existence. Secondly, we use the 3D semantic encoder trained from the previous step to generate 3D semantic representation and extract the features of the linguistic instructions by the pre-trained BERT model. Finally, we use an LSTM-based navigation model with attention layers to incorporate the language feature and 3D semantic representations for the VLN task. In this figure, Q, K and V indicate the inputs of the Multi-head cross-modal attention network.}
    \label{framework}
    \end{figure*}
% \cite{bert}

Our main contributions are summarized as follows:

\begin{itemize}
    % We introduce the self-supervised 3D semantic representation for the vision-and-language navigation task.
    \item We analyze the important role of 3D semantic representation for the vision-and-language navigation task and establish a new learning and fusion framework.
    \item We utilize the region query pretext task to design a self-supervised learning method for 3D semantic representation from unlabelled 3D semantic reconstructions.
    \item  Experimental results in Room2Room(R2R) dataset show that our proposed method brings more than 10\% absolute improvement in the success rate compared with the previous RGB-D method\cite{sun2021depth}. In addition, with an LSTM-based navigation model, our proposed method outperforms most of RGB-based methods utilizing vision-language transformers.
\end{itemize}

\section{Related Work}

\noindent \textbf{Methods for the VLN Task}
Recently, there are many methods proposed for the VLN task \cite{vln}, which mainly focuses on improving the cross-modal vision-language association and scene generalization capabilities. For vision-language reasoning,
% need to change 
\cite{smna,RCM,aux} have introduced cross-modal attention structures and additional tasks such as navigation progress and path matching degree predictions to establish the correspondence between observations and instructions over each step. For scene generalization, the data augmentation methods for instructions and scenarios are investigated in \cite{sf,envdrop,envmixup}. 

In addition, along with the widespread use of pre-trained transformers for vision-language pairing tasks,% \cite{vlbert,lxmert,vilbert}, 
vision-language transformers are introduced to the VLN task to model cross-modal reasoning at each step \cite{prevalent,orist,recurrent}. However, most of the above methods only use RGB observation features and fail to effectively integrate the depth information into the navigation tasks.

\noindent \textbf{Depth and 3D Semantic Representation for Embodied Navigation}
%\noindent \textbf{3D semantic representation for embodied navigation}
The 3D semantic information provides valuable clues for the embodied agent to navigate in the environment. It can be obtained by using the depth sensors and has already been used in the visual semantic navigation task \cite{chaplot2020object}, where an image or a semantic label is denoted as the target for the agent to navigate to, and thus a powerful language processing capability is not necessarily required. On the other hand, in the language-rich visual navigation tasks such as vision-and-language navigation task, the 3D semantic representation 
with depth information are rarely used,% which is partially due to the complicated coupling between 3D features and the language representation.
which is partially due to the difficulty of training deep models with 3D semantic inputs in an end-to-end approach.
\cite{sun2021depth} attempts to employ the depth features into the navigation model, while only marginal improvement is observed, implying the difficulty of incorporating the 3D information into the vision-and-language navigation task.

\noindent \textbf{Self-supervised Learning}
Self-supervised feature learning aims to learn effective feature representation from large amounts of the unlabelled data, which can then be used for downstream tasks. It has been extensively investigated in computer vision tasks.
%  \cite{he2020momentum,chen2020simple}
Recently, there have been some works attempting to apply the self-supervised representation learning technique to embodied tasks. For example, a self-supervised learning framework called SEAL\cite{chaplot2021seal} is proposed, which utilizes perception models trained on internet images to learn an active exploration policy. In our work, we adopt self-supervised learning for 3D semantic representation in the vision-and-language navigation task.

\section{Method}

In this section, we will present the learning framework of the 3D semantic representation constructed from RGB-D inputs for navigation tasks as shown in Fig.\ref{framework}. Firstly, we describe the encoding procedure of how observation inputs are converted to 3D semantic representations. In the second and third parts, we introduce encoding approaches of linguistic instructions and how the 3D features are adapted to the vision-and-language navigation task. Finally, we discuss the fusion strategy, which leverages the advantages of both RGB and 3D semantic features.

\subsection{Self-Supervised Learning for 3D Semantic Representation}
Here we first introduce the encoder for the 3D semantic representation, and then present the proposed self-supervised learning method.

\subsubsection{3D Semantic Reconstruction} %TODO??

To reconstruct the 3D environment from the RGB-D observations, we use Swin Transformer trained on ADE20k with 150 classes, though other segmentors could be utilized. We first feed multiple 2D RGB observations of an agent at a specific location into the segmentor to obtain 2D semantic segmentations, which are then projected into 3D space using depth observation and camera parameters. After then, each RGB-D observation is converted into a 3D semantic point cloud. By combining the 3D point clouds from different viewpoints, we can obtain a location-specific panoramic semantic point cloud. Please note that the used voxelization parameters are $0.125$ m for the $X$ and $Y$ axes and $0.25$m for the $Z$-axis. For the agent, the maximum range of horizon observation with respect to the center is $\pm 8$m, and for vertical is $\pm 4$m. As a result, the dimension of the voxel-based 3D semantic reconstruction is $150 \times 128 \times 128 \times 32$, which leads to much memory consumption for the navigation training.

To tackle this problem, we use a sparse representation of the 3D reconstruction, only recording locations with at least one non-zero semantic category. This will produce a tensor of shape $k \times 4$ recording the index in a mini-batch and the locations, and a tensor of shape $k \times 150$ to record the semantic categories, where $k$ is the number of locations with a non-zero semantic category. Finally, the sparse convolution network is used to encode the sparse representations of 3D reconstruction input to a $2048 \times 4 \times 4$ tensor, which is denoted as $\mathcal{F}_{sem}$.

% \cite{liu2021swin}\cite{ade20k}

%  \cite{sparseconv}
    
\subsubsection{Self-Supervised Learning Method}
    \begin{figure*}[htbp]
    \centering
    \includegraphics[width=0.9\linewidth]{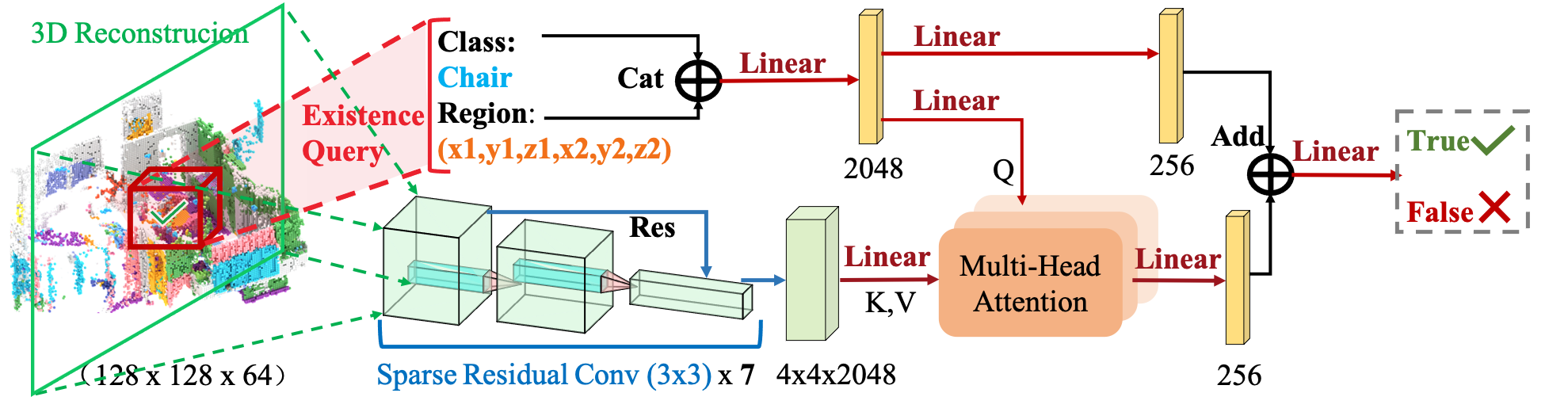}
    \caption{The model structure used for self-supervised training. The 3D semantic reconstruction is encoded by the sparse convolutional network. The visual embedding as K, V and concatenated query vectors as Q will be fed into the multi-headed attention network, respectively. As indicated by the red box in the figure, the self-supervised training task requires the model to recognize the existence of a certain object class in the specified region.}
    \label{fig:pretrain_model}
    \end{figure*}
   Though we design the encoder for the 3D semantic representation, how to determine the massive network parameters is challenging due to the limited training samples in the VLN task.

    To address this problem, we develop a self-supervised approach to pre-train sparse convolution networks to encode the semantic reconstruction. To enable the network to deal with the 3D reconstruction, we design a \textit{region query} pretext task\footnote{The final accuracy of our proposed region query task for the 3D reconstructions is 96.5\% on the training set, and 95.4\% on the validation set.}, in which the 3D visual encoding model is required to answer object presence questions in specific regions. The task is defined as the following:

    Given a query, which is defined as a tuple $(x_1, y_1, z_1, x_2, y_2, z_2, c)$, asking whether there is an object of class $c$ in the region satisfying $x_1 \leq x \leq x_2$, $y_1 \leq y \leq y_2$ and $z_1 \leq z \leq z_2$, the model should give the answer \textit{True} or \textit{False}. This is a two-class classification problem and its training samples and labels can be obtained in the concerned scenarios.

    In the following we introduce how to use this pretext task to train the network. First, the query is encoded with a Multi-Layer Perception (MLP) to a concatenated 2048-$d$ vector $\mathbf{q}$, which is represented as
    \begin{equation}
    \begin{split}
    \mathbf{q} &= {MLP}([(x_1, x_2, y_1, y_2, z_1, z_2); W_c \mathbf{c}]) \\
    \end{split}
    \end{equation}
    where $\mathbf{c}$ is the one-hot vector for the class $c$ and $W_c$ is the learnable parameter. Then we use the softmax output of the multi-head attention-based model to estimate the probability of the answer $ans\in \{True, False\}$ as
    \begin{equation}
    prob({ans}) = {softmax}(W_h \mathbf{h} + W_{q} \mathbf{q})
    \end{equation}
    where $W_h$, $W_{q}$ are learnable parameters, and $\mathbf{h}$ is defined as the output of the multi-head attention layer:
    \begin{equation}
    \mathbf{h} = {Attn}(\mathbf{q}, K, V)
    \end{equation}
    where ${Attn}$ denotes the standard multi-head attention layer, and both $K$ and $V$ are set to the 3D semantic feature $\mathcal{F}_{sem}$. The model architecture for the self-supervised training task is illustrated in Fig.\ref{fig:pretrain_model}.
    % \cite{vaswani2017attention}

    To use this pretext task for training, we need to prepare training data, which requires a set of regions with object existence annotation. Since randomly sampled regions in the semantic reconstruction are likely to contain much more negative samples than positive samples for the region query task, we use a simple balancing technique to modify the training samples: For a mini-batch of 3D semantic reconstructions, we first randomly sample many regions containing at least one kind of object in one reconstruction. Then we count the number of positive and negative samples for each class and truncate the number of negative samples to be the same as the number of positive samples. To further improve the generalization ability of the model trained in the proposed self-supervised approach, we apply random noise and affine transformations to the point cloud, including scaling, translation, and rotation to generate more various scenes as the training data for the pretext task. We expect the proposed 3D visual encoding model to perceive semantic information in different ranges of observation spaces via a self-supervised query task. The association of spatial locations in the scene with the semantic information of objects benefits downstream navigation tasks.
    
    \subsection{Language Encoder}

    In this work, we use BERT as the language encoder, which produces two groups of vectors of the navigation instruction: $f_L$ as the language feature, and $f_{L}^*$ as the sentence level feature representation. These language representations will be further used by the navigation model for language understanding.
    
    \subsection{Navigation Model}
    
        The navigation model is designed based on the LSTM-based model proposed in \cite{envdrop}. At the initial time step, the hidden state $\mathbf{h}_0$, cell state $\mathbf{c}_0$ and vision-language context representation $\mathbf{g}_0$ are initialized from the sentence level language representation $f_L^*$. At time instant $t$, it takes the current 3D semantic feature and the language feature $f_L$ as the input, and use the previous vision-language context $\mathbf{g}_t$ to query the 3D semantic visual features and fuse with it using a two-layer MLP network $\psi$ with GELU activation and residue connection to get 
    \begin{equation}
        \begin{split}
        \mathbf{v}_t = \psi({Attn}(\mathbf{g}_t, K^V, V^V), \mathbf{g}_t) \\
        \end{split}
    \end{equation}
    where $K^V$ and $V^V$ are also set to the current 3D semantic. Further, $\mathbf{v}_t$ is concatenated with the angle feature of the previous action and then forwarded through an LSTM unit used for modeling the observation sequence of the agent, getting the next cell state $\mathbf{c}_{t+1}$ and the next hidden state $\mathbf{h}_{t+1}$.

    The output hidden state of the LSTM cell is used to query both the language feature and the 3D semantic feature for further understanding of both the language context and the visual context, and the context representation is updated as 
    \begin{equation}
        \begin{split}
            Q^L &= \psi({Attn}(Q^H, K^L, V^L), Q^H) \\
            \mathbf{g}_{t+1} &= \psi({Attn}(Q^L, K^V, V^V), Q^L) \\
        \end{split}
    \end{equation}
    where $K^L$ and $V^L$ are set to $f_L$, and $Q^H$ are set to $\mathbf{h}_{t+1}$.
    Finally, the action selection is performed using a simple network with bilinear operation between $\mathbf{g}_{t+1}$ and the angle encoding of each viewport following \cite{sf}. We use both Imitation Learning (IL) and A2C-based Reinforcement Learning (RL) losses for the navigation model like \cite{RCM}, and follow \cite{recurrent} for reward shaping and the balancing factor between the IL loss and the RL loss.

    \subsection{Fusion of 3D Semantic Feature with RGB Feature}
    
As previously analyzed, RGB features have more enriched detail information than 3D semantic features. Therefore, using RGB and 3D semantic features together in the network can potentially further improve the performance of the model.

When doing so, we simultaneously perform the same attention operations to both the 3D semantic feature and RGB feature. The final results are average of them. However, such a fusion strategy may not work well since RGB network branches are trained from scratch with their parameters initialized randomly, and tend to make the model overfit due to the enriched information on texture details of RGB inputs. This will also be verified by our experiments.

%  though straightforward, 

To address this problem, we propose to use the learned 3D semantic representation to constraint the training process of the RGB branches. To verify this point, we develop a cross-modal distillation technique to initialize the RGB branch of the network. Concretely speaking, we first train the RGB network to force its output to mimic that of the 3D semantic feature network. This procedure provides a better initialization for the RGB network because it preserves more high-level structure information. After that, the RGB branch of the network are jointly trained with the 3D branch in the final finetuning stage.

    \section{Experiment and Results}
    \begin{table*}[h]
    \caption{Comparison of results on R2R validation unseen and test unseen splits in single-run setting with other methods.† works that use transformer with vision-language pre-training and models in gray shading uses depth information.}
    \label{table_experiment_results_other_sota}
    \centering
    \begin{tabular}{|c|c|c|c|c|c|c|c|c|c|}
    \hline \hline 
    \multirow{2}{*}{\textbf{Methods}} & \multicolumn{3}{|c|}{\textbf{R2R Validation Seen}} & \multicolumn{3}{|c|}{\textbf{R2R Validation Unseen}} & \multicolumn{3}{|c|}{\textbf{R2R Test Unseen}} \\
    \cline{2-10}
    & \textbf{NE}$\downarrow$ & \textbf{ SR}$\uparrow$ & \textbf{SPL}$\uparrow$ & \textbf{ NE}$\downarrow$ & \textbf{ SR}$\uparrow$ & \textbf{ SPL}$\uparrow$ & \textbf{ NE}$\downarrow$ & \textbf{  SR}$\uparrow$ & \textbf{ SPL}$\uparrow$ \\
    \hline
    % \multicolumn{10}{|c|}{\textbf{Methods using panoramic input and action space proposed in \cite{sf}}} \\
    % \hline
    Random\cite{vln} & 9.45 & 16 & - & 9.23 & 16 & - & 9.79 & 13 & 12 \\
    Human\cite{vln} & - & - & - & - & - & - & 1.61 & 86 & 76 \\
    \hline
    Speaker-Follower\cite{sf} & 3.36 & 66 & - & 6.62 & 35 & - & 6.62 & 35 & 28 \\
    RCM+SIL(train)\cite{RCM}& 3.53 & 67 & - & 6.09 & 43 & - & 6.12 & 43 & 38 \\
    % \textbf{PRESS\cite{press}}& 4.39 & 58 & 55 & 5.28 & 49 & 45 & 5.49 & 49 & 45 \\
    EnvDrop\cite{envdrop}& 3.99 & 62 & 59 & 5.22 & 52 & 48 & 5.23 & 51 & 47 \\
    % \textbf{OAAM\cite{oaam}} & - & 65 & 62 & - & 54 & 50 & - & 53 &50 \\
    AuxRN\cite{aux}& 3.33 & 70 & 67 & 5.28 & 55 & 50 & 5.15 & 55 & 51 \\
    RelGraph\cite{relgraph}&3.47 & 67 & 65 & 4.73 & 57 & 53 & 4.75 & 55 & 52 \\
    % \textbf{\thead{CMG-AAL \cite{cmgaal}}} & $\mathbf{2.74}$ & $\mathbf{73}$ & $\mathbf{69}$ & 4.18 &59 &51 & 4.61 &57 &50 \\
    % \textbf{Regretful\cite{regretful}}& 3.23 & 69 & 63 & 5.32 & 50 & 41 & 5.69 & 48 & 40 \\
    SMNA\cite{smna} & 3.22 & 67 & 58 & 5.52 & 45 & 32 & 5.67 & 48 & 35 \\
    % \textbf{FAST\cite{fast}}& - & - & - &  4.97 & 56 & 43 & 5.14 & 54 & 41 \\
    % \textbf{Active Perception\cite{active}} &3.20 &70 &52 & 4.36 &58 &40 & 4.33 &60 &41 \\
    SSM\cite{ssm} & 3.10 & 71 & 62 & 4.32  & 62 & 45 & 4.57 & 61 & 46 \\
    %\hline
    %\multicolumn{10}{|c|}{\textbf{PREVALENT-based model using vision-language transformer \cite{sf}}} \\
    %\hline
    PREVALENT \cite{prevalent}†& 3.67 & 69 & 65 & 4.71 & 58 & 53 & 5.30 & 54 & 51 \\
    \rowcolor{mygray} DASA \cite{sun2021depth}† & 3.76 &67 &64 & 4.77 &58 &54 & 5.11 &54 &52 \\
    ORIST \cite{orist}† & - &- &- & 4.72 &57 &51 & 5.10 &57 &52 \\
    \thead{VLN$\circlearrowright$BERT \cite{recurrent}†}& $2.90$ & $72$ & $68$ & $3.93$ & $63$ & $57$ & $4.09$ & $63$ & $57$ \\
    {HAMT}\cite{hamt}† & \textbf{2.51} & \textbf{76} & \textbf{72} & \textbf{2.29} & 66 & 61 & 3.93 & 65 & 60 \\
    \rowcolor{mygray} {\textbf{Ours}} & {2.55} & 74 & 70 & 3.33 & \textbf{68} & \textbf{61} & \textbf{3.73} & \textbf{66} & \textbf{60} \\

    \hline \hline
    \end{tabular}
    \end{table*}

    \begin{table*}[h]
    \caption{Ablation experiments on the effectiveness of applying various visual features. The checkmarks indicate the corresponding used features, where * denotes whether the RGB model use the initialization method in Section 3.4, in the navigation training process. † Denotes that for the final validation, we use a max action length of 30 instead of 15 for further improving the SR of the proposed model.}
    % + work 
    \label{table_ablation_analysis}
    \centering
    \begin{tabular}{|c|c|c|c|c|c|c|c|c|c|c|c|}
    \hline \hline 
     & \multicolumn{3}{|c|}{\textbf{Features}} & \multicolumn{4}{|c|}{\textbf{R2R Validation Seen}} & \multicolumn{4}{|c|}{\textbf{R2R Validation Unseen}} \\
    % \cline{2-9}
    \hline
     \textbf{Models} & \textbf{3D Semantic} & \textbf{RGB} & \textbf{RGB*} & \textbf{ TL } & \textbf{NE}$\downarrow$ & \textbf{ SR}$\uparrow$ & \textbf{SPL}$\uparrow$ & \textbf{ TL } & \textbf{ NE}$\downarrow$ & \textbf{ SR}$\uparrow$ & \textbf{ SPL}$\uparrow$ \\
    \hline
    1 &$\boldsymbol{\checkmark}$ & & & 11.0 & 2.70 & 73.1 & 69.5 & 12.1 & 3.46 & 66.8 & 60.8 \\
    2 & & $\boldsymbol{\checkmark}$ & & 11.2 & 3.07 & 69.3 & 65.9 & 12.0 & 4.31 & 58.1 & 52.6 \\
    3 & & & $\boldsymbol{\checkmark}$ & 11.5 & 3.21 & 67.9 & 63.6 & 12.0 & 3.73 & 63.3 & 57.4 \\
    % & \textcolor{red}{\boldsymbol{\checkmark}} & 11.0 & 3.99 & 62.1 & 59 & 10.7 & 5.22 & 52.2 & 48 \\
    4 &$\boldsymbol{\checkmark}$ & $\boldsymbol{\checkmark}$ & & 11.6 & 2.63 & 73.8 & 69.5 & 12.6 & 3.55 & 65.7 & 59.0 \\
    5 &$\boldsymbol{\checkmark}$ & &
    $\boldsymbol{\checkmark}$ & 11.5 & 2.58 & 73.7 & 69.8 & 12.0 & \textbf{3.33} & 67.7 & \textbf{61.4} \\
    6† & $\boldsymbol{\checkmark}$ & & $\boldsymbol{\checkmark}$ & 12.7 & \textbf{2.55} & \textbf{74.1} & \textbf{69.9} & 14.3 & \textbf{3.33} & \textbf{67.9} & 61.2 \\
    \hline \hline
    \end{tabular}
    \end{table*}

    \subsection{Experiment Setup}
    For training the 3D semantic encoder, we use the AdamW optimizer with a learning rate of 4e-5. In addition, we use 64 heads for the multi-head attention model in the region query network training. In order to extract panoramic RGB features, we use ResNet-152 pretrained on the Places365 dataset. Furthermore, to language encoder, We finetune the last encoder layer of the BERT model in the training process. 
    
    After obtaining a semantic representation of the scene with the 3D semantic encoder, we use a three-stage approach to train the network with that representation on the R2R dataset. In the first stage, we use the Speaker-Follower augmented data provided by PREVALENT for training, with ~1M instruction-trajectory pairs generated using the Speaker model with a learning rate of 4e-5. Secondly, we use the R2R standard training set for training, with a learning rate of 1e-5. In the third stage, we train the proposed 3D semantic feature model jointly with the RGB branch added. In these experiments, We use AdamW as the optimizer for the navigation agent. We always use a hidden size of 1024 for all hidden layers in the navigation models and a batch size of 256 in the training process. 
    
    \subsection{Experiment Results}
    
    \begin{figure*}
    \centering
    \includegraphics[width=0.9\linewidth]{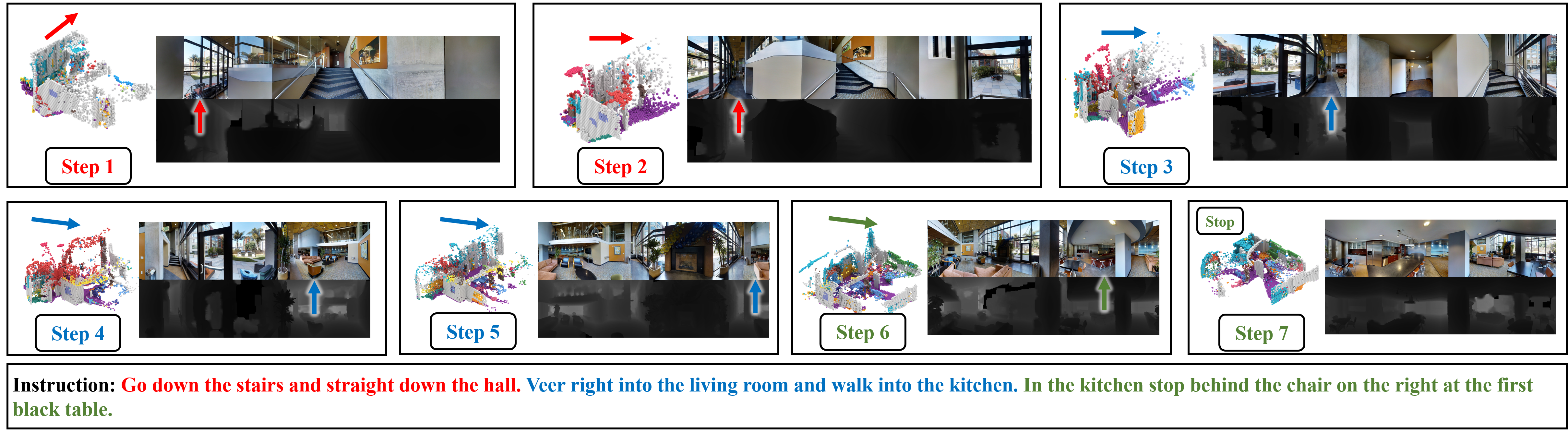}
    % where an .eps filename suffix will be assumed under latex, 
    % and a .pdf suffix will be assumed for pdflatex; or what has been declared
    % via \DeclareGraphicsExtensions.
    \caption{A qualitative example showing the actions when the agent is following the instruction ``Go down the stairs and straight down the hall. Veer right into the living room and walk into the kitchen. In the kitchen stop behind the chair on the right at the first black table.''. Arrows and texts in the same color represent related action and textual description. Differences between the panoramic image and the semantic reconstruction feature space are also depicted.}
    \label{fig:qualitative}
    \end{figure*}
We submit our experiment results to the public evaluation server\footnote{\url{https://eval.ai/web/challenges/challenge-page/97} Method name: 3DSR} to evaluate the performance on the R2R dataset. The results are summarized in Table \ref{table_experiment_results_other_sota}, from which we get the following key observations.

Taking advantage of the 3D semantic reconstruction obtained from the RGB-D observation, the proposed method outperforms most of the published RGB-based methods on the R2R dataset in the single-run setup achieving a success rate of 66\% on the test unseen splits of the R2R dataset.

It is noted that we use an LSTM-based navigation model rather than the large-scale pre-trained vision-language transformer structure used by PREVALENT \cite{prevalent}, Recurrent VLN-BERT \cite{vln-bert} and HAMT\cite{hamt}. Nevertheless, our method still has improvements in the success rates when compared with them.

Also, compared to the recently developed DASA model \cite{sun2021depth} which also utilizes the depth information, our proposed method brings a more than 10\% absolute improvement of the success rate on the test unseen split. This indicates that the proposed encoding and learning method for 3D semantic representation is more effective than the existing method using depth information.

\begin{figure}
    \centering
    
    \includegraphics[width=0.8\linewidth]{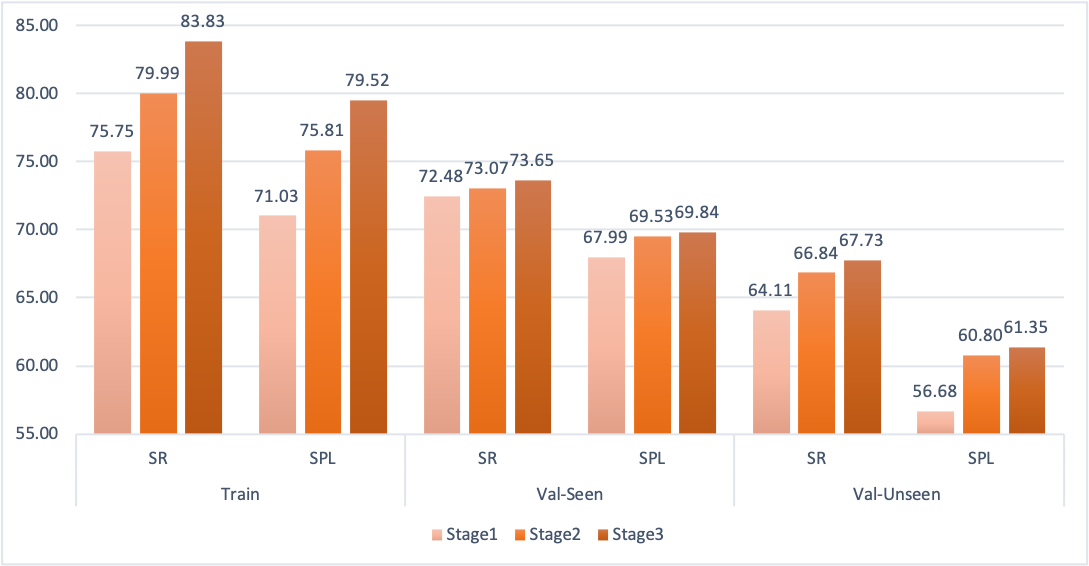}
    \caption{Performance of the navigation model on three different training stages. In stage one, the navigation model was trained with the augmented data generated by the speaker-follower model and was trained only using the R2R dataset in the other two stages. In the first two stages, only the 3D semantic presentation was applied, and in stage three, we introduced the RGB-2D branches with the initialization method in Section 3.4 to the model structure to complement the 3D features for further performance improvements.}
    \label{fig:diffstage}
    \end{figure}

In Fig.\ref{fig:diffstage}, we show the influences of different training stages for the proposed method. With the first-stage training on the Speaker-augmented dataset, a success rate of 64.1\% is obtained, which shows the effectiveness of the Speaker-Follower data augmentation method \cite{prevalent}. In the second stage, the model is finetuned on the original train split of the R2R dataset. This further improves the performance of the model to 66.8\%. Finally, we finetune the proposed model with the RGB branch initialized techniques introduced in Section 3.4 and a success rate of 67.7\% is achieved.

\subsection{Ablation Analysis}

\subsubsection{Is 3D Semantic Feature Better than RGB Feature?}
To compare the proposed features with the standard RGB panoramic ResNet features, which are commonly used in the R2R task. We replace the 3D semantic representation in our proposed training framework with typical panoramic RGB features and the experimental results are reported in model 2 of Table 2. The results show that utilizing RGB features leads to a success rate of 58.1\%, which is significantly lower than 66.8\% when using the proposed 3D semantic features, demonstrating the advantages of the developed 3D semantic features for the vision-and-language navigation task.

\subsubsection{Does 3D Semantic Feature Improve RGB Feature Learning?}
% \subsubsection{Impact of RGB branch model initialization}

We firstly test the performance of the proposed model using the RGB feature, which is trained from scratch, and get a success rate of 58.1\% on the validation unseen split. Further, we initialize the RGB branch of the model by forcing the RGB branch to mimic the output of the 3D semantic branch of the network, and the feature is denoted as RGB$^*$. As model 2 and model 3 of Table 2, the performance can be improved to 63.3\%, which is comparable with most RGB-based models. This result indicates that the 3D semantic feature can be used to improve network structures with other input modalities in the training process.
 
\subsubsection{Does the Fusion of 3D Semantic Feature and RGB Feature Work?}
% Complementarity between RGB feature and 3D reconstruction feature
When we use the fusion of 3D semantic feature and RGB feature which the RGB branch is initialized from scratch, the model can only achieve a success rate of 65.7\% on the validation unseen split, which is worse than the result when training with the 3D semantic feature alone as model 1 and model 4 of Table 2. However, if we use the fusion of 3D semantic feature and RGB$^*$ feature which the RGB model branch use the initialization method as Section 3.4, the proposed model can be further finetuned and achieve a success rate of 67.7\%, which is better than the result of using the 3D semantic feature only (66.8\%) and the result of only using initialized RGB network branch (63.3\%) on the validation unseen split. This shows that the RGB and 3D semantic features may be complementary to each other, and therefore, it is possible to use them together to further improve the performance of the single modal framework.

\subsection{Qualitative Result}
Fig.\ref{fig:qualitative} shows a qualitative example of the VLN task using the proposed model trained on the R2R dataset. The 3D voxel reconstruction in each step demonstrates the input of the 3D semantic encoder. All these 3D semantic reconstructions are generated from panoramic RGB-D observation of the sensors. The colored arrow's direction indicates the movement direction of the agent.

\section{Conclusions and Future Works}

In this work, we propose a self-supervised representation learning framework to build 3D semantic representations from RGB-D observations, and use it in the vision-and-language navigation task. The encoding network for local 3D semantic reconstruction is firstly trained with a region query pretext task in a self-supervised manner. And then, the extracted 3D semantic features are fed into an LSTM-based navigation model for the vision-and-language navigation task. Experimental results show that our proposed method outperforms existing methods with and achieves success rates of 68\% and 66\% on the validation unseen and test unseen splits of the R2R dataset respectively. For future works, we would like to discuss the possibility of replacing the depth information we used in this work with some depth prediction or multi-view reconstruction methods.

    % \addtolength{\textheight}{-12cm} 
    
    %\section*{ACKNOWLEDGMENT}
    
    %This work was supported in part by the National Natural Science Foundation of China under Grants 62025304.

    %%%%%%%%%%%%%%%%%%%%%%%%%%%%%%%%%%%%%%%%%%%%%%%%%%%%%%%%%%%%%%%%%%%%%%%%%%%%%%%%
    
\bibliographystyle{named}
\bibliography{ijcai22}

\begin{thebibliography}{}

\bibitem[\protect\citeauthoryear{Anderson \bgroup \em et al.\egroup
  }{2018}]{vln}
Peter Anderson, Qi~Wu, Damien Teney, Jake Bruce, Mark Johnson, Niko
  S{\"u}nderhauf, Ian Reid, Stephen Gould, and Anton Van Den~Hengel.
\newblock Vision-and-language navigation: Interpreting visually-grounded
  navigation instructions in real environments.
\newblock In {\em Proceedings of the IEEE Conference on Computer Vision and
  Pattern Recognition}, pages 3674--3683, 2018.

\bibitem[\protect\citeauthoryear{Chaplot \bgroup \em et al.\egroup
  }{2020}]{chaplot2020object}
Devendra~Singh Chaplot, Dhiraj~Prakashchand Gandhi, Abhinav Gupta, and Russ~R
  Salakhutdinov.
\newblock Object goal navigation using goal-oriented semantic exploration.
\newblock {\em Advances in Neural Information Processing Systems}, 33, 2020.

\bibitem[\protect\citeauthoryear{Chaplot \bgroup \em et al.\egroup
  }{2021}]{chaplot2021seal}
Devendra~Singh Chaplot, Murtaza Dalal, Saurabh Gupta, Jitendra Malik, and
  Russ~R Salakhutdinov.
\newblock Seal: Self-supervised embodied active learning using exploration and
  3d consistency.
\newblock {\em Advances in Neural Information Processing Systems}, 34, 2021.

\bibitem[\protect\citeauthoryear{Chen \bgroup \em et al.\egroup }{2021}]{hamt}
Shizhe Chen, Pierre-Louis Guhur, Cordelia Schmid, and Ivan Laptev.
\newblock History aware multimodal transformer for vision-and-language
  navigation.
\newblock {\em Advances in Neural Information Processing Systems}, 34, 2021.

\bibitem[\protect\citeauthoryear{Fried \bgroup \em et al.\egroup }{2018}]{sf}
Daniel Fried, Ronghang Hu, Volkan Cirik, Anna Rohrbach, Jacob Andreas,
  Louis-Philippe Morency, Taylor Berg-Kirkpatrick, Kate Saenko, Dan Klein, and
  Trevor Darrell.
\newblock Speaker-follower models for vision-and-language navigation.
\newblock In {\em Proceedings of the 32nd International Conference on Neural
  Information Processing Systems}, pages 3318--3329, 2018.

\bibitem[\protect\citeauthoryear{Hao \bgroup \em et al.\egroup
  }{2020}]{prevalent}
Weituo Hao, Chunyuan Li, Xiujun Li, Lawrence Carin, and Jianfeng Gao.
\newblock Towards learning a generic agent for vision-and-language navigation
  via pre-training.
\newblock In {\em 2020 {IEEE/CVF} Conference on Computer Vision and Pattern
  Recognition, {CVPR} 2020, Seattle, WA, USA, June 13-19, 2020}, pages
  13134--13143. Computer Vision Foundation / {IEEE}, 2020.

\bibitem[\protect\citeauthoryear{Hong \bgroup \em et al.\egroup
  }{2020}]{relgraph}
Yicong Hong, Cristian Rodriguez, Yuankai Qi, Qi~Wu, and Stephen Gould.
\newblock Language and visual entity relationship graph for agent navigation.
\newblock {\em Advances in Neural Information Processing Systems},
  33:7685--7696, 2020.

\bibitem[\protect\citeauthoryear{Hong \bgroup \em et al.\egroup
  }{2021}]{recurrent}
Yicong Hong, Qi~Wu, Yuankai Qi, Cristian~Rodriguez Opazo, and Stephen Gould.
\newblock {VLN} {BERT:} {A} recurrent vision-and-language {BERT} for
  navigation.
\newblock In {\em {IEEE} Conference on Computer Vision and Pattern Recognition,
  {CVPR} 2021, virtual, June 19-25, 2021}, pages 1643--1653. Computer Vision
  Foundation / {IEEE}, 2021.

\bibitem[\protect\citeauthoryear{Liu \bgroup \em et al.\egroup
  }{2021}]{envmixup}
Chong Liu, Fengda Zhu, Xiaojun Chang, Xiaodan Liang, Zongyuan Ge, and Yi-Dong
  Shen.
\newblock Vision-language navigation with random environmental mixup.
\newblock In {\em Proceedings of the IEEE/CVF International Conference on
  Computer Vision}, pages 1644--1654, 2021.

\bibitem[\protect\citeauthoryear{Ma \bgroup \em et al.\egroup }{2019}]{smna}
Chih-Yao Ma, Jiasen Lu, Zuxuan Wu, Ghassan AlRegib, Zsolt Kira, Richard Socher,
  and Caiming Xiong.
\newblock Self-monitoring navigation agent via auxiliary progress estimation.
\newblock {\em arXiv preprint arXiv:1901.03035}, 2019.

\bibitem[\protect\citeauthoryear{Majumdar \bgroup \em et al.\egroup
  }{2020}]{vln-bert}
Arjun Majumdar, Ayush Shrivastava, Stefan Lee, Peter Anderson, Devi Parikh, and
  Dhruv Batra.
\newblock Improving vision-and-language navigation with image-text pairs from
  the web.
\newblock In {\em European Conference on Computer Vision}, pages 259--274.
  Springer, 2020.

\bibitem[\protect\citeauthoryear{Qi \bgroup \em et al.\egroup }{2020}]{reverie}
Yuankai Qi, Qi~Wu, Peter Anderson, Xin Wang, William~Yang Wang, Chunhua Shen,
  and Anton van~den Hengel.
\newblock Reverie: Remote embodied visual referring expression in real indoor
  environments.
\newblock In {\em Proceedings of the IEEE/CVF Conference on Computer Vision and
  Pattern Recognition}, pages 9982--9991, 2020.

\bibitem[\protect\citeauthoryear{Qi \bgroup \em et al.\egroup }{2021}]{orist}
Yuankai Qi, Zizheng Pan, Yicong Hong, Ming{-}Hsuan Yang, Anton van~den Hengel,
  and Qi~Wu.
\newblock The road to know-where: An object-and-room informed sequential bert
  for indoor vision-language navigation.
\newblock In {\em ICCV}, pages 1655--1664, 2021.

\bibitem[\protect\citeauthoryear{Sun \bgroup \em et al.\egroup
  }{2021}]{sun2021depth}
Qiang Sun, Yifeng Zhuang, Zhengqing Chen, Yanwei Fu, and Xiangyang Xue.
\newblock Depth-guided adain and shift attention network for
  vision-and-language navigation.
\newblock In {\em 2021 IEEE International Conference on Multimedia and Expo
  (ICME)}, pages 1--6. IEEE, 2021.

\bibitem[\protect\citeauthoryear{Tan \bgroup \em et al.\egroup
  }{2019}]{envdrop}
Hao Tan, Licheng Yu, and Mohit Bansal.
\newblock Learning to navigate unseen environments: Back translation with
  environmental dropout.
\newblock In {\em Proceedings of the 2019 Conference of the North American
  Chapter of the Association for Computational Linguistics: Human Language
  Technologies, Volume 1 (Long and Short Papers)}, pages 2610--2621, 2019.

\bibitem[\protect\citeauthoryear{Thomason \bgroup \em et al.\egroup
  }{2020}]{cvdn}
Jesse Thomason, Michael Murray, Maya Cakmak, and Luke Zettlemoyer.
\newblock Vision-and-dialog navigation.
\newblock In {\em Conference on Robot Learning}, pages 394--406. PMLR, 2020.

\bibitem[\protect\citeauthoryear{Wang \bgroup \em et al.\egroup }{2019}]{RCM}
Xin Wang, Qiuyuan Huang, Asli Celikyilmaz, Jianfeng Gao, Dinghan Shen,
  Yuan{-}Fang Wang, William~Yang Wang, and Lei Zhang.
\newblock Reinforced cross-modal matching and self-supervised imitation
  learning for vision-language navigation.
\newblock In {\em {IEEE} Conference on Computer Vision and Pattern Recognition,
  {CVPR} 2019, Long Beach, CA, USA, June 16-20, 2019}, pages 6629--6638.
  Computer Vision Foundation / {IEEE}, 2019.

\bibitem[\protect\citeauthoryear{Wang \bgroup \em et al.\egroup }{2021}]{ssm}
Hanqing Wang, Wenguan Wang, Wei Liang, Caiming Xiong, and Jianbing Shen.
\newblock Structured scene memory for vision-language navigation.
\newblock In {\em Proceedings of the IEEE/CVF Conference on Computer Vision and
  Pattern Recognition}, pages 8455--8464, 2021.

\bibitem[\protect\citeauthoryear{Zhu \bgroup \em et al.\egroup }{2020}]{aux}
Fengda Zhu, Yi~Zhu, Xiaojun Chang, and Xiaodan Liang.
\newblock Vision-language navigation with self-supervised auxiliary reasoning
  tasks.
\newblock In {\em Proceedings of the IEEE/CVF Conference on Computer Vision and
  Pattern Recognition}, pages 10012--10022, 2020.

\end{thebibliography}
    
    \end{document}